\documentclass{article}

\usepackage{microtype}
\usepackage{graphicx}
\usepackage{subcaption}
\usepackage{booktabs} 

\usepackage{hyperref}

\usepackage[preprint]{icml2026}

\usepackage{amsmath}
\usepackage{amssymb}
\usepackage{mathtools}
\usepackage {amsthm}
\usepackage{multirow}

\usepackage[capitalize,noabbrev]{cleveref}

\theoremstyle{plain}

\theoremstyle{definition}

\theoremstyle{remark}

\usepackage[textsize=tiny]{todonotes}
\usepackage{color}

\usepackage{booktabs}
\usepackage{multirow}
\usepackage[table]{xcolor} 
\usepackage{amsmath}       

\usepackage{graphicx}
\usepackage{subcaption}

\definecolor{rowblue}{RGB}{236, 244, 252} 
\definecolor{mygreen}{RGB}{46, 139, 87}   
\newcommand{\g}[1]{\textcolor{mygreen}{#1}} 

\begin{document}

\twocolumn[
  \icmltitle{Reversible Diffusion Decoding for Diffusion Language Models}



  \icmlsetsymbol{equal}{*}
  \icmlsetsymbol{co}{\dag}
  \icmlsetsymbol{pl}{\ddag}

  \begin{icmlauthorlist}
    \icmlauthor{Xinyun Wang}{ecnu}
    \icmlauthor{Min Zhang}{co,pl,ecnu}
    \icmlauthor{Sen Cui}{thu}
    \icmlauthor{Zhikang Chen}{oxford}
    \icmlauthor{Bo Jiang}{ecnu}
    \icmlauthor{Kun Kuang}{zju}
    \icmlauthor{Mingbao Lin}{rakuten}
  \end{icmlauthorlist}

  \icmlaffiliation{ecnu}{East China Normal University}
  \icmlaffiliation{thu}{Tsinghua University}
  \icmlaffiliation{zju}{Zhejiang University}
  \icmlaffiliation{rakuten}{Rakuten Singapore}
  \icmlaffiliation{oxford}{University of Oxford}

  \icmlcorrespondingauthor{Min Zhang}{mzhang@cs.ecnu.edu.cn}

  \icmlkeywords{Machine Learning, ICML}

  \vskip 0.3in
]



\printAffiliationsAndNotice{\icmlCorrespondingauthor \icmlProjectLeader}


\newcommand{\METHOD}{RDD}

\begin{abstract}
    Diffusion language models enable parallel token generation through block-wise decoding, but their irreversible commitments can lead to stagnation, where the reverse diffusion process fails to make further progress under a suboptimal context.
    We propose Reversible Diffusion Decoding (RDD), a decoding framework that introduces reversibility into block-wise diffusion generation. RDD detects stagnation as a state-dependent failure of the reverse process and enables efficient backtracking to earlier blocks without recomputation via cached model states. To avoid repeated failure trajectories, RDD applies confidence-guided re-masking to selectively reinitialize uncertain tokens while preserving reliable context.
    This reversible formulation allows decoding to recover from early commitment errors while maintaining the parallel efficiency of diffusion-based generation. Experiments show that RDD improves generation robustness and quality over baselines with minimal computational overhead.
\end{abstract}

\section{Introduction}

Diffusion language models (DLMs) have recently emerged as a promising alternative to the conventional paradigm of autoregressive (AR)  generation by enabling parallel token updates through bidirectional context modeling~\cite{dream,llada}.
By iteratively denoising masked tokens, DLMs generate text in a non-causal manner, alleviating inherent limitations of left-to-right decoding such as exposure bias and the reversal curse~\cite{welleck2019non,berglundreversal}.
Empirical results have demonstrated that diffusion-based generation can achieve competitive quality compared to AR models while offering new opportunities for parallelism during inference.

Despite their conceptual appeal, practical deployment of DLMs remains challenging.
In contrast to highly optimized AR systems, diffusion-based generation often incurs substantially higher inference cost. 
In particular, naive parallel decoding degrades generation quality, while the absence of a straightforward key-value (KV) caching mechanism limits efficiency gains from modern hardware.
These issues have motivated a growing body of work on accelerating DLM inference. The typical studies often include block-wise semi-autoregressive decoding~\cite{blockdiffusion,fastdllm,saber,d2f}, cache-aware formulations~\cite{dllmcache,fastdllm,d2cache}, and speculative or auxiliary decoding strategies~\cite{ebsampler,selfspecdllm}.

A common design principle underlying these acceleration techniques is \textbf{monotonic generation}: once a block of tokens is generated, it is irrevocably committed to the context and treated as fixed for all subsequent steps. 
While this assumption simplifies scheduling and caching, it introduces a critical vulnerability.
In discrete text generation, locally plausible decisions made in early blocks may later prove globally suboptimal, particularly for long-range dependencies, structured reasoning, or syntactic consistency.
When later blocks are conditioned on such suboptimal prefixes, the reverse diffusion process may fail to confidently denoise any remaining tokens, causing generation to stall.
We refer to this state-dependent failure mode as \textbf{stagnation}.

Existing approaches address stagnation only indirectly. Forced decoding heuristics relax confidence thresholds to ensure progress, often amplifying hallucination errors~\cite{d2f}. Self-correction methods iteratively refine low-confidence tokens during inference or through additional training~\cite{saber,remdm,r3,remedi}. Although effective in certain regimes, these strategies typically incur substantial test-time overhead, require specialized models, or operate primarily at the token level. More importantly, they do not directly address the block-level error propagation induced by accelerated diffusion decoding, where irreversible commitments prevent recovery from early mistakes.

In this work, we propose \textbf{Reversible Diffusion Decoding (\METHOD)}, a decoding framework that introduces \emph{reversibility} into block-wise diffusion generation. Instead of enforcing monotonic commitments,~\METHOD~treats stagnation as a state-dependent failure of the reverse diffusion process and enables efficient
rollback to earlier blocks when progress becomes impossible under the current context. 
\METHOD~performs such rollback without recomputation and selectively reinitializes uncertain tokens through confidence-guided re-masking, allowing the generation process to escape suboptimal trajectories.

Crucially,~\METHOD~is designed to complement rather than replace existing acceleration techniques. It operates entirely at inference time, requires no additional training or auxiliary models, and preserves the efficiency benefits of block-wise decoding. Building on this reversible formulation, we further introduce an adaptive scheduling strategy that dynamically adjusts denoising granularity based on the decoding state, enabling aggressive parallel generation
on easy segments while allocating additional computation only when recovery is necessary.

Empirical results demonstrate that~\METHOD~substantially improves generation robustness and quality over standard block diffusion decoding, while incurring minimal computational overhead. By addressing a fundamental limitation of accelerated diffusion inference---irreversible block-level commitments---\METHOD~
provides a principled and practical pathway toward efficient and reliable diffusion-based language generation.

The major contributions we have made in this paper include:
\begin{itemize}
    \item We identify \textbf{stagnation} as a state-dependent failure mode of accelerated diffusion decoding arising from irreversible block-level commitments.
    \item We propose \textbf{Reversible Diffusion Decoding (\METHOD)}, a decoding framework that enables efficient rollback and selective re-initialization at block boundaries without retraining or recomputation.
    \item We introduce an adaptive scheduling strategy layered on top of~\METHOD~that improves efficiency while preserving robustness.
    \item Extensive experiments show that~\METHOD~improves generation quality and robustness across diffusion language models with minimal overhead.
\end{itemize}
\section{Related Work}

\subsection{Acceleration of Diffusion Language Models}

Diffusion Language Models (DLMs) depart from conventional autoregressive (AR) generation by enabling parallel token updates through bidirectional context modeling, alleviating inherent limitations such as the reversal curse. Representative models including Dream~\cite{dream} and LLaDA~\cite{llada} operate directly in discrete token space and perform denoising via bidirectional attention. While this paradigm allows for parallel decoding and holistic context modeling, it introduces new challenges in inference efficiency. In particular, the absence of an explicit key–value (KV) cache and quality degradation under aggressive parallel sampling cause DLMs to lag behind highly optimized AR models in practice.

In order to narrow down this factual efficiency gap, most of the existing studies have explored acceleration along three main directions including: cache-aware decoding, parallel generation strategies, and speculative or auxiliary decoding. Block Diffusion~\cite{blockdiffusion}, Fast-dLLM~\cite{fastdllm}, and d$^2$Cache~\cite{d2cache} adopt semi-autoregressive formulations that enable KV caching and block-wise parallel decoding. EB-Sampler~\cite{ebsampler} improves the quality of parallel sampling, while D2F~\cite{d2f} accelerates inference through block-level causal attention and pipeline parallelism. APD~\cite{apd} further incorporates auxiliary autoregressive models to assist fast decoding.

Though the above advances have been made, existing acceleration techniques share a common structural assumption: \textbf{generation proceeds monotonically}. Once a block is generated, it is treated as immutable context and cannot be revised. However, errors introduced by early local decisions therefore propagate forward, leading to error accumulation and degraded global coherence. This limitation fundamentally constrains the robustness of accelerated DLM inference and motivates the need for decoding mechanisms that can revise prior commitments.

\subsection{Self-Correction Mechanisms in DLMs}


To enhance the generation quality of DLMs, recent studies have introduced self-correction mechanisms that can be broadly categorized into training-free and training-based approaches.
Training-free methods typically identify uncertain or low-confidence tokens at inference time and iteratively refine them. Saber~\cite{saber} introduces backtracking-enhanced remasking guided by historical confidence scores. ReMDM~\cite{remdm} performs selective remasking based on token-level uncertainty, while R3~\cite{r3} proposes a “Review, Remask, and Refine” framework driven by a process-level reward model to enable targeted corrections. In contrast, training-based approaches explicitly learn correction behaviors. ReMeDi~\cite{remedi}, for example, trains specialized models under the UPS framework to identify erroneous tokens and perform refinement.

\begin{figure*}[t]
    \centering
    \begin{subfigure}[t]{0.48\textwidth}
        \centering
        \includegraphics[width=\linewidth]{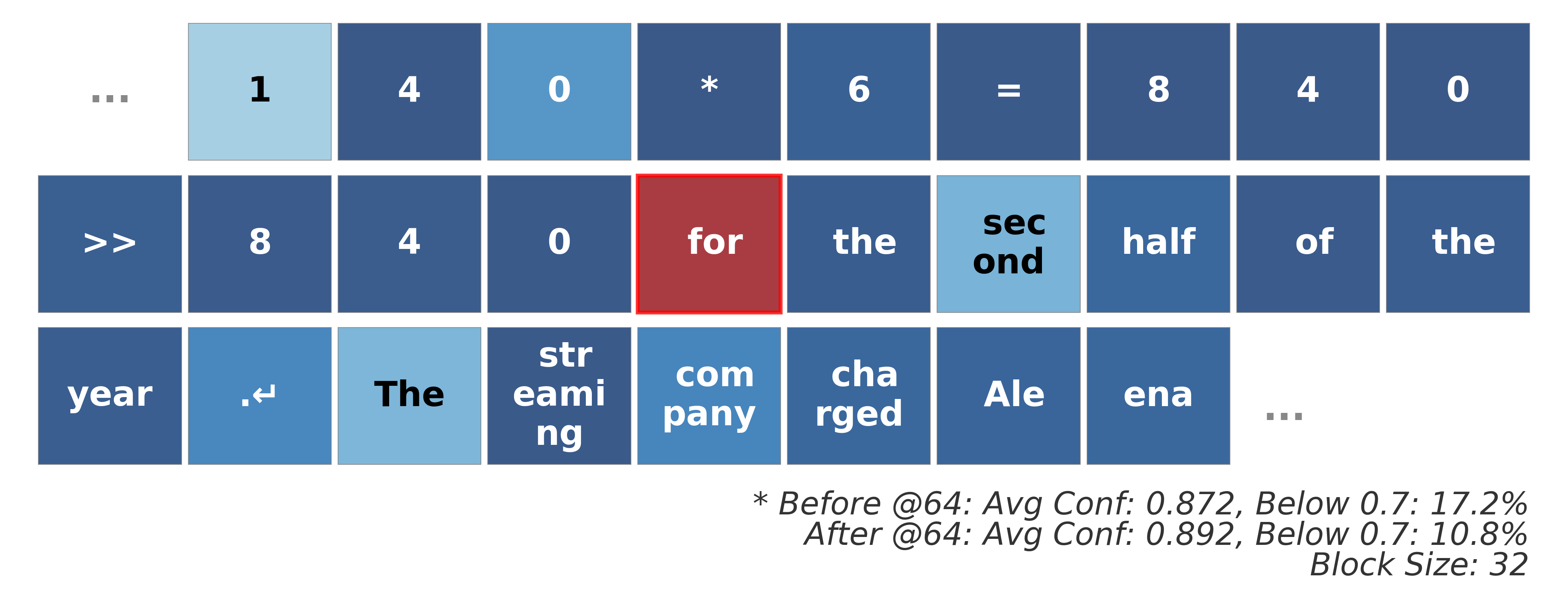}
        \caption{}
        \label{fig:stagnation_fastdllm_rdd}
    \end{subfigure}
    \hfill
    \begin{subfigure}[t]{0.48\textwidth}
        \centering
        \includegraphics[width=\linewidth]{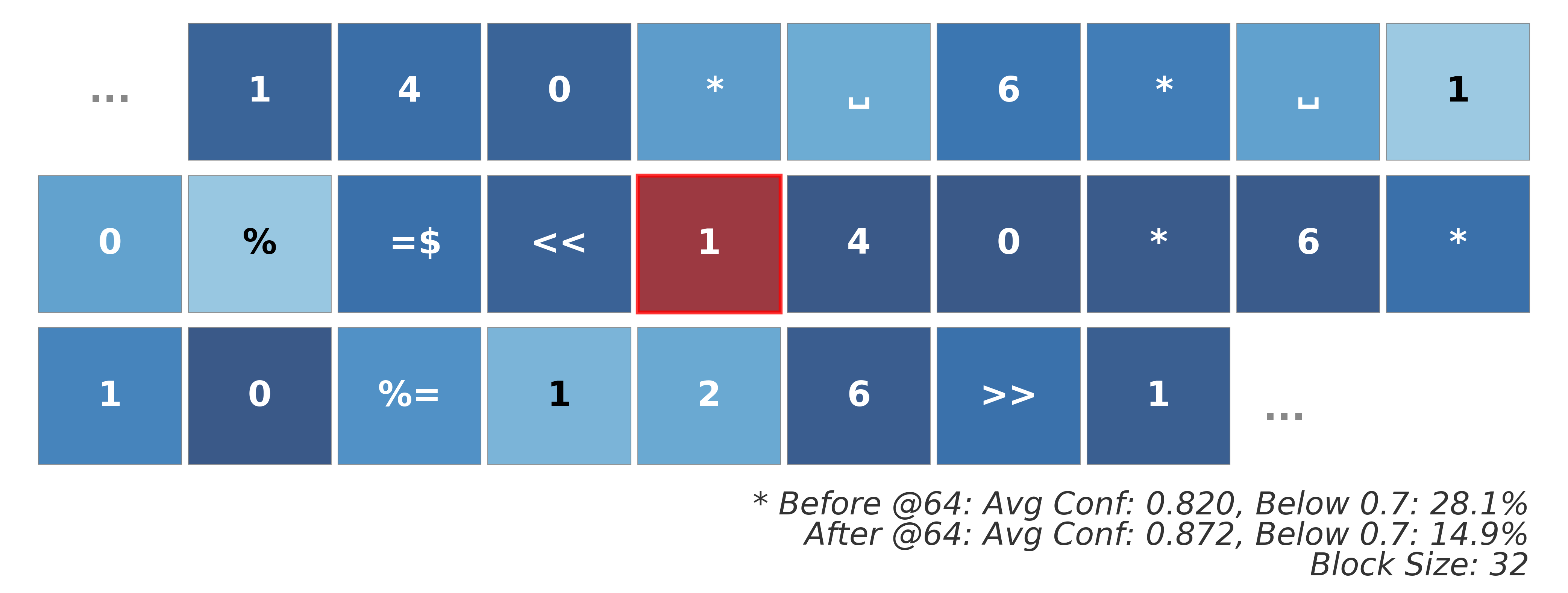}
        \caption{}
        \label{fig:stagnation_rdd_fastdllm}
    \end{subfigure}
    \caption{
        \textbf{Visualizing the Stagnation Trap and the Impact of Prefix Quality.} 
        We compare generation trajectories by switching decoding strategies at token 64 ({\color{red}red token}). Token color indicates prediction confidence, darker is better.
        \textbf{(a)} Decoding the first 64 tokens with \METHOD~yields a high-confidence prefix (Avg Conf: 0.872). This robust context allows the standard Fast-dLLM to maintain high stability in the suffix (Avg Conf: 0.892).
        \textbf{(b)} Decoding the prefix with Fast-dLLM results in less certain (Avg Conf: 0.820). Crucially, because these suboptimal tokens are locked in the context, even switching to the robust \METHOD~cannot fully recover the generation (Avg Conf: 0.872), which remains less confident than the method in (a). Further details are provided in Appendix~\ref{app:stagnation-trap}.
    }
    \label{fig:stagnation_trap}
    \vspace{-9pt} 
\end{figure*}

While effective at improving quality, these methods differ from acceleration-oriented decoding. Training-free approaches often rely on extensive test-time scaling or increased memory usage, which can significantly slow inference. Training-based methods~\cite{d2f, remedi} incur substantial training cost and require specialized models, limiting their flexibility and ease of deployment. More importantly, existing self-correction techniques operate at the token level, failing to explicitly address \textbf{block-level error propagation} induced by semi-autoregressive generation. Furthermore, token-level self-correction cannot effectively leverage the KV cache from previous blocks, resulting in \textbf{significant computational overhead and diminished inference speed}.

\section{Methodology}

We now present the methodological foundation of our approach. This section formalizes the decoding setting for diffusion language models and analyzes the failure modes induced by block-wise acceleration. Building on this analysis, we introduce a reversible decoding framework that enables principled recovery from stagnation while remaining compatible with existing cache-aware and semi-autoregressive decoding schemes. The following subsections detail the problem formulation, the proposed reversible mechanism, and its efficiency-oriented extensions.

\subsection{Problem Setup and Background}

\textbf{Diffusion Language Models.} 
Let $X_0 = (x_{0,1}, \dots, x_{0,L})$ be a clean token sequence. 
Diffusion language models (DLMs) define a discrete forward corruption process over an extended vocabulary $\mathcal{V} \cup \{[\text{MASK}]\}$, where tokens are progressively masked according to a predefined noise schedule.
At diffusion step $t$, each token is independently masked with probability
\begin{equation}
    q(x_{t,i} = [\text{MASK}] | x_{0,i}) = 1 - \bar{\alpha}_t,
\end{equation}
where $\bar{\alpha}_t \in [0, 1]$ monotonically decreases with $t$.

The reverse process is modeled by a neural network $p_\theta(X_{t-1} | X_t)$, which predicts the original token distribution for masked positions.
For each masked token $i$, the model outputs
\begin{equation}
    \hat{p}_i = p_\theta(x_{0,i} | X_t) \in \Delta^{|\mathcal{V}|}.
\end{equation}

We define the \textbf{prediction confidence} at position $i$ as
\begin{equation}
    p_{conf}(x_{0,i}) = \max_{v \in \mathcal{V}}\hat{p}_i(v),
\end{equation}
which measures the model’s certainty about the reconstructed token.

\textbf{Block-wise Semi-Autoregressive Decoding.}
To accelerate generation, block-wise diffusion decoding~\cite{blockdiffusion} partitions the sequence into $K$ contiguous blocks $\mathbf{B} = \{B_1, \dots, B_K\}$, each of length $m$.
Blocks are generated sequentially but decoded internally in parallel:
\begin{equation}
    B_k \sim p_\theta(B_k | \mathbf{C}_k), \; \mathbf{C}_k = [B_1, ..., B_{k-1}].
\end{equation}

This semi-autoregressive formulation improves efficiency by parallelizing intra-block generation while preserving global coherence through prefix conditioning.

\subsection{Failure Mode: The Stagnation Trap}

Despite its efficiency, standard block diffusion introduces a critical rigidity: \textbf{once a block is finalized, it is irrevocably committed to the context}.
In discrete text generation, locally plausible decisions in early blocks may later prove globally suboptimal, particularly for long-range dependencies or reasoning chains.

When subsequent blocks are conditioned on such suboptimal prefixes, the reverse diffusion process may fail to confidently unmask any remaining tokens. The generation then enters a low-confidence state in which no further progress is possible without violating decoding constraints.
We refer to this state-dependent failure mode as the \textbf{Stagnation Trap}, show as \cref{fig:stagnation_trap}.

Existing approaches~\cite{fastdllm,d2f} mitigate stagnation via forced decoding heuristics (\emph{e.g.}, top-$k$ fallback), which prematurely lock in uncertain predictions and often amplify hallucination errors.
We argue that robustness in diffusion decoding fundamentally requires \textbf{reversibility}---the ability to retract and revise earlier commitments when they hinder future generation.

\subsection{Reversible Block Diffusion Decoding}
We address the \textit{stagnation trap} by introducing Reversible Diffusion Decoding (\METHOD), a decoding framework that enables the reverse diffusion process to retract and revise earlier block-level commitments.~\METHOD~formalizes stagnation as a state-dependent failure of the reverse process and recovers from it through controlled rollback and selective re-initialization.

\textbf{Decoding State.}
Consider block $B_k$ at reverse diffusion step $s$. We denote the current decoding state as
\begin{equation}
    \mathcal{S}_{k, s} = \big( X_{k, s}, \; \mathbf{C}_k, \; 
    \mathcal{K}_k\big),
\end{equation}
where $X_{k,s}$ is the partially denoised block, $\mathbf{C}_k = [B_1,\ldots,B_{k-1}]$ is the prefix context, and $\mathcal{K}_k$
denotes the cached key–value state.

Let
\begin{equation}
\mathcal{M}_s = \{\, i \mid x^{(s)}_{k,i} = [\mathrm{MASK}] \,\}
\end{equation}
be the set of remaining masked positions at step $s$.

\textbf{Decodable Token Set.}
Given a confidence threshold $\tau(\lvert \mathcal{M}_s\rvert)$, the set of confidently decodable tokens is
\begin{equation}
\mathcal{U}_s(\tau) =
\left\{
i \in \mathcal{M}_s \;\middle|\;
p_{\mathrm{conf}}(x_{k,i}) \ge \tau(\lvert \mathcal{M}_s\rvert)
\right\}.
\end{equation}

\textbf{Stagnation Detection.}
We define a decoding state if stagnant if
\begin{equation}
\mathcal{U}_s(\tau) = \emptyset
\quad \text{and} \quad
\lvert \mathcal{M}_s \rvert > 0.
\end{equation}

This condition indicates that the reverse diffusion process cannot make further progress under the current context and noise schedule.

\textbf{Rollback Operator.}
To recover from stagnation,~\METHOD~introduces a block-level rollback operator

\begin{equation}
\mathcal{R}_r : \mathcal{S}_{k,s} \mapsto \mathcal{S}_{k-r,n},
\end{equation}

which reverts decoding to the earlier state of block $B_{k-r}$.
Then, the rolled-back blocks will be merged into the generating block.

\begin{equation}
    \hat{B}_{k} = B_{k-r\dots k}.
\end{equation}


\textbf{State Transition Rule.}
The decoding process follows the transition
\begin{equation}
\mathcal{S}_{k,s+1} =
\begin{cases}
\textsc{Decode}(\mathcal{S}_{k,s}), & \mathcal{U}_s(\tau) \neq \emptyset, \\
\mathcal{R}_r(\mathcal{S}_{k,s}), & \mathcal{U}_s(\tau) = \emptyset,
\end{cases}
\end{equation}
where $r$ is increased up to a maximum depth $r_{\max}$. 
If stagnation persists after $r_{\max}$ rollbacks, forced decoding is applied to guarantee termination.

\textbf{Confidence-Guided Re-masking.}
After rollback, each token $x_{k, i}$ in the revived block is re-masked independently with probability
\begin{equation}
p_{\mathrm{remask}}(x_{k,i}) =
1 - p_{\mathrm{conf}}(x_{k,i})^{\lambda},
\end{equation}
where $\lambda$ controls sensitivity to uncertainty. This preserves high-confidence structure while encouraging exploration over ambiguous regions.

The full procedure of \METHOD~is detailed in~\cref{alg:main_method}.

\begin{algorithm}[htbp]
\caption{\METHOD~Algorithm}
\begin{algorithmic}[1]
\REQUIRE DLM $p_\theta$, Context $C$, Block Length $L$, Scaling $f$ and $f_r$, Rollback Budget $R$, Remask sensitivity $\lambda$, Total Length $L_t$
\STATE $X \leftarrow C \cup \{[\text{MASK}] \dots\}$

\FOR{$b_s \leftarrow |C|$ to $L_t$ step $L$}
    \STATE $\text{budget} \leftarrow R$, $f_{curr} \leftarrow f$, $b_{e} \leftarrow b_{s}+L$
    \STATE $x_{b_s}, \text{Cache} \leftarrow p_\theta(B_{curr})$
    \WHILE{$\exists$ [MASK] in $B_{curr}$}
        \STATE $B_{curr} \leftarrow X[b_{s} : b_{e}]$
        \STATE $P, \text{Cache} \leftarrow p_\theta(B_{curr} | \text{Cache})$
            \STATE $\tau \leftarrow 1 - f / (\text{count\_mask}(B_{curr}) + 1)$
        \STATE $\mathcal{I} \leftarrow \{i \mid P_i > \tau \text{ and } x_i = \text{[MASK]}\}$
        \IF{$\mathcal{I} \neq \emptyset$}
            \STATE Fill $X[\mathcal{I}]$ with predicted tokens
        \ELSIF{$\text{budget} > 0$ and $b_{s} > |C|$}
            \STATE $\text{budget} \leftarrow \text{budget} - 1$, $b_{s} \leftarrow b_{s} - L$
            \STATE $X[b_{s} : b_{e}-L] \leftarrow \text{Remask}(X, b_{s}, b_{e}-L)$
            \STATE $\text{Cache} \leftarrow \text{Cache\_Delete}(b_{s}, b_{e})$
            \STATE \textbf{break} to outer loop
        \ELSE
            \STATE $i^* \leftarrow \text{argmax}(\text{P})$, $x_{i^*} \leftarrow \text{top token}$ 
        \ENDIF
    \ENDWHILE
\ENDFOR
\label{alg:main_method}
\end{algorithmic}
\end{algorithm}

\subsection{Efficiency Optimization via Adaptive Dual-Scale Scheduling}

\begin{figure*}[t]
  \begin{center}
    \centerline{\includegraphics[width=\textwidth]{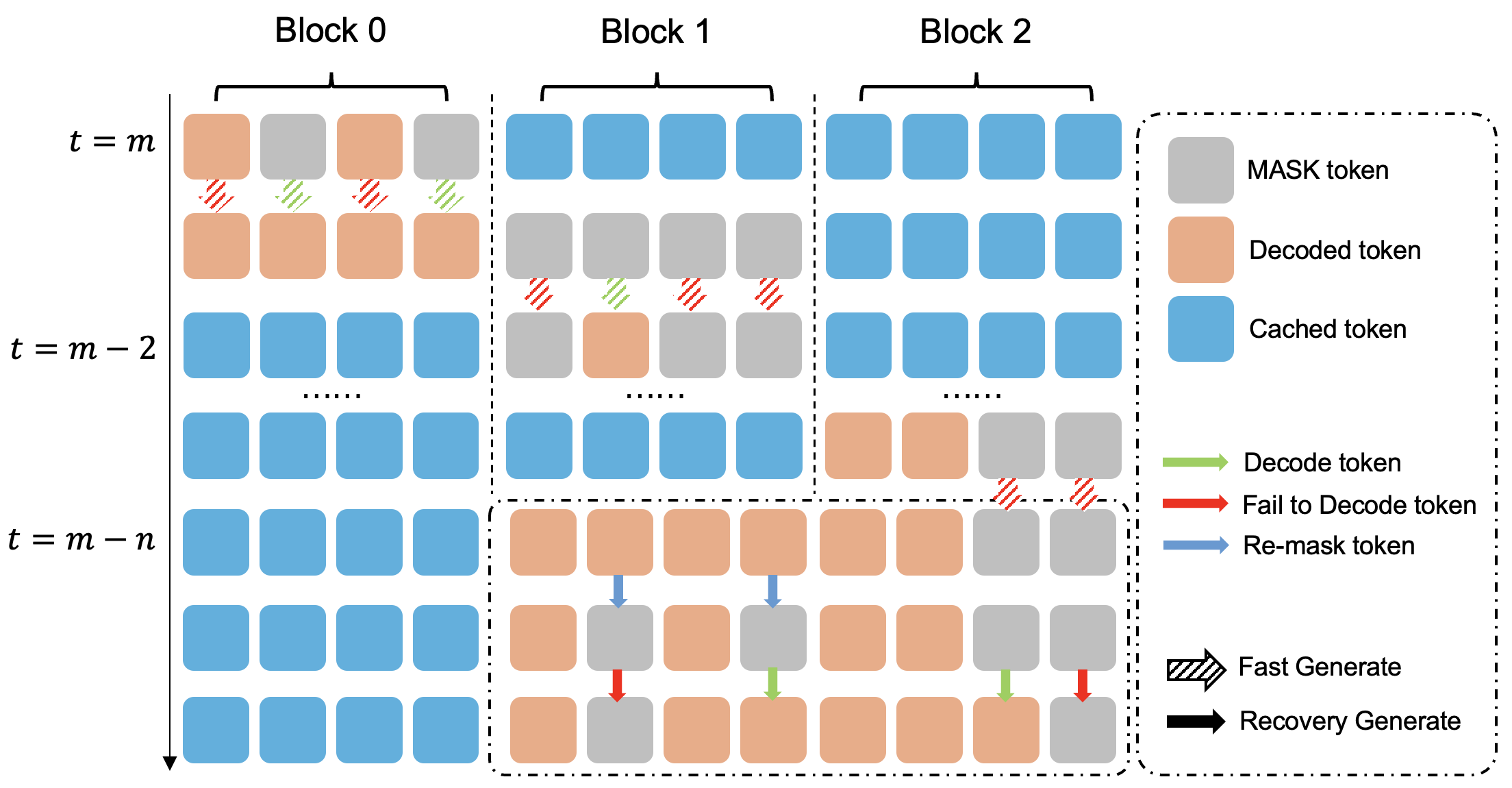}}
    \caption{
      \textbf{Illustration of the \METHOD~Process.} 
      The diagram depicts the decoding flow across three blocks. 
      \textbf{Block 0} demonstrates successful parallel generation. 
      In \textbf{Block 2}, the model falls into the \textit{Stagnation Trap} at the final denoising steps, where the reverse diffusion process fails to confidently unmask the remaining tokens. 
      Instead of forcing a suboptimal commitment, \METHOD~triggers a \textbf{Rollback} operation. 
      As shown in \textbf{Block 1}, the system efficiently retrieves the previous state and applies confidence-guided \textbf{Re-masking} to the uncertain regions, enabling the model to correct the error and resume valid generation.
    }
    \label{fig:main-figure}
    \vspace{-25pt}
  \end{center}
\end{figure*}
While~\METHOD~enables robust recovery from stagnation, rollback introduces additional computational overhead. We further introduce Adaptive Dual-Scale Scheduling for an efficiency optimization layered on top of~\METHOD~.

We interpret stagnation as an endogenous signal of increased generation difficulty. The denoising schedule is used to control exploration granularity: aggressive schedules favor speed by unmasking many tokens in parallel, while conservative schedules improve resolution by restricting updates to fewer confident predictions.

Given the number of remaining masked tokens, we define
\begin{equation}
\tau(\lvert \mathcal{M}_s\rvert) = 1 - \frac{f}{\tau(\lvert \mathcal{M}_s\rvert) + 1},
\end{equation}
where the scaling factor $f$ controls the aggressiveness of parallel decoding.
This formulation encourages large parallel updates when uncertainty is low and
automatically tightens as unresolved ambiguity accumulates.

\textbf{Dual-Scale Policy.}
We instantiate this principle using two decoding modes:

\textit{Fast-Paced Normal Mode.}
When decoding proceeds without stagnation, we adopt a large scaling factor
$f$, enabling aggressive parallel unmasking. Since the majority of text
segments fall into this regime, overall throughput remains high.

\textit{Conservative Recovery Mode.}
When stagnation is detected and re-masking is triggered, the schedule switches to a
conservative factor $f_r \ll f$. This reduces the size of $\mathcal{U}_s(\tau)$,
forcing the model to resolve uncertainty incrementally and perform a
fine-grained search over ambiguous regions.

\textbf{Efficiency---Robustness Trade-off.}
By conditioning the decoding schedule on an internal failure signal rather than
a fixed heuristic, Adaptive Dual-Scale Decoding approximates an optimal
compute-allocation strategy: it runs fast by default and slows down only when
the model explicitly indicates difficulty. This allows \METHOD~to improve
robustness without sacrificing efficiency on easy segments.

The complete workflow integrating \METHOD~with Adaptive Dual-Scale Decoding is visualized in \cref{fig:main-figure}.

\section{Experiments}

We conduct extensive experiments to systematically evaluate the effectiveness, robustness, and efficiency of the proposed Reversible Diffusion Decoding framework. Our evaluation spans multiple diffusion language models and diverse math-reasoning and code-generation benchmarks, with a particular focus on generation quality under accelerated decoding. We compare against representative baseline methods and perform ablation studies to isolate the contribution of each component. Unless otherwise specified, experimental settings are kept consistent across methods to ensure fair comparison.

\begin{table*}[t]
    \centering
    \caption{Comprehensive evaluation results on Dream-Base~\cite{dream} and LLaDA-Inst~\cite{llada}. The highest accuracy is \textbf{bold}, the second highest is \underline{underlined}, and \textcolor{mygreen}{green texts} denote the speedup ratios.}
    \label{tab:main_experiment}
    
    \begin{tabular}{ll ccc ccc}
    \toprule
    \multirow{2}{*}{\textbf{Dataset}} & \multirow{2}{*}{\textbf{Method}} & \multicolumn{3}{c}{\textbf{Dream-Base}} & \multicolumn{3}{c}{\textbf{LLaDA-Inst}} \\
    \cmidrule(lr){3-5} \cmidrule(lr){6-8}
     & & \textbf{Throughput} $\uparrow$ & \textbf{Latency(s)} $\downarrow$ & \textbf{Score} $\uparrow$ & \textbf{Throughput} $\uparrow$ & \textbf{Latency(s)} $\downarrow$ & \textbf{Score} $\uparrow$ \\
    \midrule

    \multirow{4}{*}{\shortstack[l]{\textbf{GSM8K} \\ \textit{5-shot} \\ \scriptsize Gen. Len. = 256}} 
      & Vanilla       & 7.07 \g{(1.0$\times$)} & 36.06 & 75.28 &  5.21 \g{(1.0$\times$)} & 43.61 & \textbf{79.23} \\
      & + Saber         & 13.71 \g{(1.9$\times$)}   & 18.59 & 68.01 & 14.88
      \g{(2.9$\times$)} & 15.66 & 57.09 \\
      & + dLLM-Cache  &  7.22 \g{(1.0$\times$)} & 35.44 & 75.89 &  10.20 \g{(2.0$\times$)} & 25.26 & 76.65 \\
      & + Fast-dLLM   & \textbf{43.97} \g{(6.2$\times$)} & \textbf{5.97} & 74.07 &  40.16 \g{(7.7$\times$)} & 5.73 & 75.82 \\
    \rowcolor{rowblue}
      & \METHOD    & 38.94 \g{(5.5$\times$)} & 6.56 & \textbf{76.88} & \underline{41.44} \g{(8.0$\times$)} & \underline{5.65} & \underline{78.70} \\
    \rowcolor{rowblue}
      & \METHOD$^\ast$    & \underline{42.06} \g{(5.9$\times$)} & \underline{6.12} & \underline{75.89} &  \textbf{48.02} \g{(9.2$\times$)} & \textbf{4.84} & 77.26 \\
    \midrule

    \multirow{4}{*}{\shortstack[l]{\textbf{MATH} \\ \textit{4-shot} \\ \scriptsize Gen. Len. = 256}} 
      & Vanilla       & 9.05 \g{(1.0$\times$)} & 28.26 & \textbf{41.46} &  7.55 \g{(1.0$\times$)} & 32.82 & \underline{38.40} \\
      & + Saber         & 29.93 \g{(3.3$\times$)}   & 8.55 & 36.78 & 19.49
      \g{(2.6$\times$)} & 13.17 & 27.26 \\
      & + dLLM-Cache  &  25.67 \g{(1.3$\times$)} & 19.94 & 40.10 &  20.69 \g{(2.7$\times$)} & 12.37 & 38.56 \\
      & + Fast-dLLM   & \textbf{57.77} \g{(6.4$\times$)} & \textbf{4.45} & 39.44 &  \textbf{63.42} \g{(8.4$\times$)} & \textbf{4.00} & 35.92 \\
    \rowcolor{rowblue}
      & \METHOD    &  50.44 \g{(5.6$\times$)} & 5.05 & 41.06 &  39.74 \g{(5.3$\times$)} & 6.26 & \textbf{39.28} \\
    \rowcolor{rowblue}
      & \METHOD$^\ast$    &  \underline{53.77} \g{(5.9$\times$)} & \underline{4.78} & \underline{41.20} &  \underline{45.28} \g{(6.0$\times$)} & \underline{5.49} & 38.26 \\
    \midrule

    \multirow{4}{*}{\shortstack[l]{\textbf{HumanEval} \\ \textit{0-shot} \\ \scriptsize Gen. Len. = 256}} 
      & Vanilla       & 19.80 \g{(1.0$\times$)} & 12.89 & \textbf{49.39} &  16.69 \g{(1.0$\times$)} & 16.83 & \underline{39.63} \\
      & + Saber         & 48.81 \g{(2.5$\times$)}   & 5.22 & 43.90 & 43.49
      \g{(2.6$\times$)} & 5.31 & 39.02 \\
      & + dLLM-Cache  &  32.73 \g{(1.7$\times$)} & 7.82 & 32.32 &  18.83 \g{(1.1$\times$)} & 13.60 & \textbf{40.24} \\
      & + Fast-dLLM   & \textbf{55.21} \g{(2.8$\times$)} & \textbf{4.62} & 47.56 &  \textbf{64.92} \g{(3.9$\times$)} & \textbf{3.78} & 32.32 \\
    \rowcolor{rowblue}
      & \METHOD    &  49.45 \g{(2.5$\times$)} & 5.31 & 48.17 &  44.51 \g{(2.7$\times$)} & 5.53 & 34.15 \\
    \rowcolor{rowblue}
      & \METHOD$^\ast$    & \underline{51.22} \g{(2.6$\times$)} & \underline{4.95} & \underline{48.78} &  \underline{50.10} \g{(3.0$\times$)} & \underline{4.86} & 30.49 \\
    \midrule

    \multirow{4}{*}{\shortstack[l]{\textbf{MBPP} \\ \textit{3-shot} \\ \scriptsize Gen. Len. = 256}} 
      & Vanilla       &  8.87 \g{(1.0$\times$)} & 28.86 & \textbf{57.40} &  5.83 \g{(1.0$\times$)} & 31.07 & \textbf{41.60} \\
      & + Saber         & 27.73 \g{(3.1$\times$)}   & 9.23 & 52.80 & 21.02
      \g{(2.4$\times$)} & 11.42 & 40.20 \\
      & + dLLM-Cache  &  25.13 \g{(2.8$\times$)} & 10.19 & 54.60 &  22.48 \g{(3.9$\times$)} & 10.95 & 38.40 \\
      & + Fast-dLLM   &  \textbf{61.29} \g{(6.9$\times$)} & \textbf{4.18} & 53.40 &  \textbf{52.97} \g{(9.1$\times$)} & \textbf{3.45} & 35.40 \\
    \rowcolor{rowblue}
      & \METHOD    &  53.52 \g{(6.0$\times$)} & 4.78 & \underline{55.20} &  38.11 \g{(6.5$\times$)} & 4.14 & \underline{41.00} \\
    \rowcolor{rowblue}
      & \METHOD$^\ast$    &  \underline{58.00} \g{(6.5$\times$)} & \underline{4.47} & 53.60 &  \underline{43.95} \g{(7.5$\times$)} & \underline{4.02} & 37.00 \\
    \midrule

    \multirow{4}{*}{\textbf{AVG}} 
      & Vanilla       & 11.20 \g{(1.0$\times$)} & 26.52 & \textbf{55.88} &  8.82 \g{(1.0$\times$)} & 31.08 & \textbf{49.72} \\
      & + Saber         & 30.04 \g{(2.7$\times$)}   & 10.40
      & 50.37 & 24.72
      \g{(2.8$\times$)} & 11.39 & 40.89 \\
      & + dLLM-Cache  &  22.69 \g{(2.0$\times$)} & 18.35 & 50.73 &  18.05 \g{(2.0$\times$)} & 15.55 & \underline{48.46} \\
      & + Fast-dLLM   &  \textbf{54.56} \g{(4.9$\times$)} & \textbf{4.80} & 53.62 &  \textbf{55.37} \g{(6.3$\times$)} & \textbf{4.24} & 44.87 \\
    \rowcolor{rowblue}
      & \METHOD    &  48.09 \g{(4.3$\times$)} & 5.43 & \underline{55.33} &  40.95 \g{(4.6$\times$)} & 5.40 & 48.28 \\
    \rowcolor{rowblue}
      & \METHOD$^\ast$    &  \underline{51.26} \g{(4.6$\times$)} & \underline{5.08} & 54.87 &  \underline{46.84} \g{(5.3$\times$)} & \underline{4.80} & 45.75 \\
    \bottomrule
    \end{tabular}
    \vspace{-9pt}
\end{table*}
\begin{figure*}[t]
    \centering
    \begin{subfigure}[t]{0.49\textwidth}
        \centering
        \includegraphics[width=\linewidth]{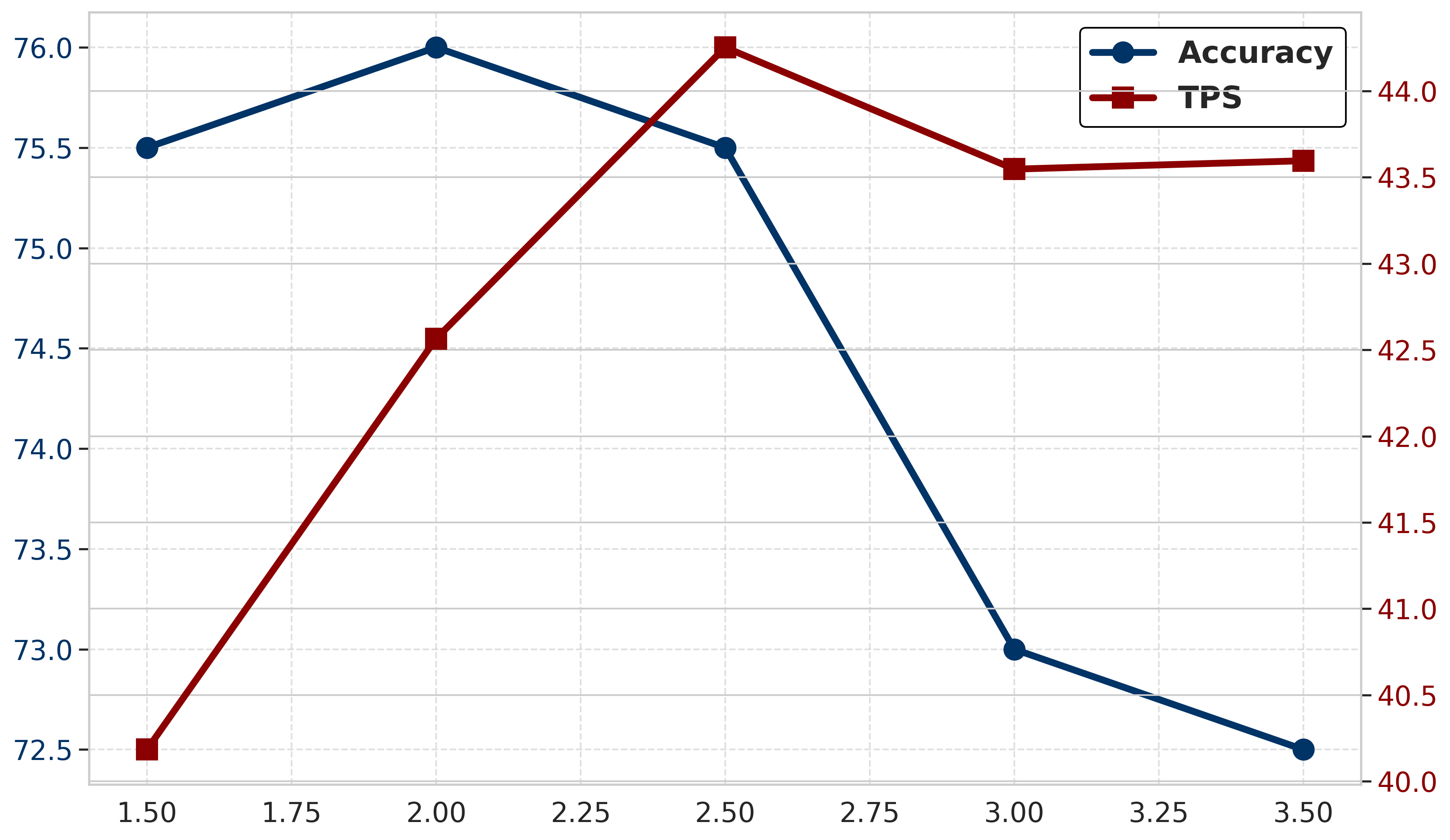}
        \caption{Performance using different $f$  for \METHOD$^\ast$~($f_r=f/2.5$)}
        \label{fig:ablation_f}
    \end{subfigure}
    \hfill
    \begin{subfigure}[t]{0.49\textwidth}
        \centering
        \includegraphics[width=\linewidth]{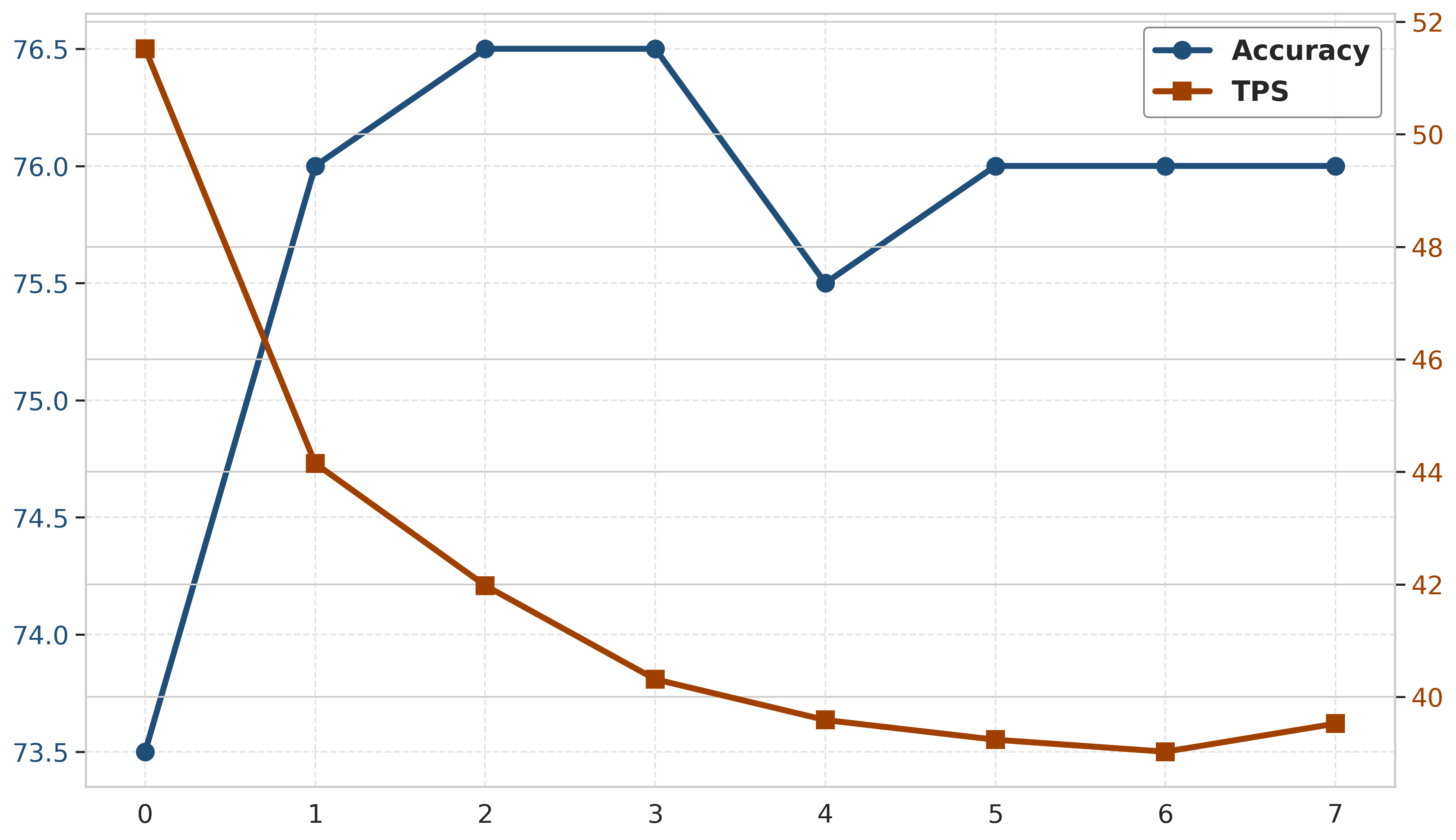}
        \caption{Performance using different $R$ for \METHOD$^\ast$}
        \label{fig:ablation_R}
    \end{subfigure}
    \caption{
        Ablation studies on key RDD hyperparameters evaluated on the GSM8K dataset.
        (a) The impact of varying the decoding scaling factor $f$ on generation accuracy and throughput.
        (b) The impact of varying the rollback budget $R$ on generation accuracy and throughput.
    }
    \label{fig:ablation_studies}
    \vspace{-9pt}
\end{figure*}

\subsection{Experimental Setup}

\textbf{Datasets.}
We consider a diverse set of reasoning and code-generation benchmarks to assess
both generation quality and robustness under accelerated decoding:
\textbf{GSM8K}~\cite{gsm8k} and \textbf{MATH}~\cite{hendrycksmath2021} for mathematical reasoning,
as well as \textbf{HumanEval}~\cite{humaneval} and \textbf{MBPP}~\cite{mbpp} for code
generation.

\textbf{Baselines.}
We compare the following decoding strategies:

(i) \textbf{Vanilla} diffusion decoding without acceleration;
(ii) \textbf{Saber}~\cite{saber}, a training-free sampling algorithm for DLMs to achieve better inference speed and output quality;
(iii) \textbf{dLLM-Cache}~\cite{dllmcache}, an
adaptive caching framework for diffusion language models;
(iv) \textbf{Fast-dLLM}~\cite{fastdllm}, a representative block-wise
semi-autoregressive acceleration method.

\textbf{Implementation details.}
For fair comparison, we set block size to 32 and the maximum generation length to 256 for all settings, and we use identical sampling hyperparameters for all of them. Dual-cache decoding is enabled for both Fast-dLLM, \METHOD~and \METHOD$^\ast$. All experiments were conducted on NVIDIA RTX 4090 GPUs.

For \METHOD~and Fast-dLLM, we set the scaling factor $f=0.9$ for Dream-Base and $f=1.0$ for LLaDA-Inst. We set remask sensitivity $\lambda=1$ and rollback budget $R=1$ (unless varied in ablations).

We additionally report a variant \textbf{\METHOD$^\ast$}, which enables the adaptive dual-scale scheduling.
For this variant, we configure the scaling and reset factors as $f=2.25, f_r=0.9$ for Dream-Base, and $f=4.0, f_r=1.0$ for LLaDA-Inst.
Unless otherwise specified,~\METHOD~refers to the base reversible decoding framework without this optimization.

\textbf{Models.} 
We evaluate our method on two representative diffusion language models: \textbf{Dream-Base}~\cite{dream} and \textbf{LLaDA-Inst}~\cite{llada}. Both models
operate in the discrete token space and perform bidirectional denoising, making
them suitable testbeds for block-wise diffusion decoding and cache-aware
acceleration.

\textbf{Metrics.}
We report three complementary metrics:
\textbf{Throughput} (tokens per second) to measure decoding efficiency,
\textbf{Latency} (seconds per sample) to reflect end-to-end inference cost, and
task-specific \textbf{Score} to evaluate generation quality (accuracy for
GSM8K and MATH, and pass@1 for HumanEval and MBPP).
All reported throughputs are measured as tokens/sec averaged over the full dataset runs; latency is measured end-to-end including cache operations.

\subsection{Main Results}

As shown in~\cref{tab:main_experiment}, the introduction of the rollback mechanism yields significant performance improvements across all tasks relative to Fast-dLLM, even surpassing Vanilla performance on specific models and datasets. Furthermore, it consistently outperforms Vanilla, Saber, and dLLM-Cache in terms of speed; notably, on the GSM8K dataset using LLaDA-Inst, \METHOD$^\ast$ even exceeds the speed of Fast-dLLM.

\textbf{Generation Quality and Robustness} 
The primary objective of \METHOD is to mitigate the quality degradation inherent in monotonic accelerated decoding. As presented in \cref{tab:main_experiment}, standard acceleration methods like Fast-dLLM frequently suffer from performance drops compared to Vanilla decoding, particularly on reasoning-intensive tasks. For example, on the MATH benchmark using Dream-Base, Fast-dLLM causes accuracy to decline from the Vanilla baseline of 41.46\% to 39.44\%. This illustrates the severity of the Stagnation Trap, where irreversible errors accumulate in the context.

\METHOD~effectively reverses this degradation. On the same MATH benchmark, \METHOD~restores performance to 41.06\% and the adaptive variant (\METHOD$^\ast$) reaches 41.20\%, effectively recovering the capabilities of the unaccelerated baseline. Notably, on GSM8K with Dream-Base, \METHOD~achieves 76.88\%, surpassing both Fast-dLLM (74.07\%) and the Vanilla baseline (75.28\%). Averaged across all tasks for Dream-Base, \METHOD~achieves a score of 55.33\%, significantly outperforming Fast-dLLM (53.62\%) and nearly matching the Vanilla model's 55.88\%. These results empirically confirm that enabling reversibility allows the model to correct early commitment errors than would otherwise lead to stagnant, low-quality generation.

\textbf{Inference Efficiency}
Despite the overhead introduced by the rollback mechanism, \METHOD~maintains high inference efficiency comparable to monotonic acceleration methods. On Dream-Base, \METHOD~achieves an average throughput of 48.09 tokens/s, representing a $4.3\times$ speedup over Vanilla decoding. While this is slightly lower than Fast-dLLM ($4.9\times$), the trade-off yields substantial gains in generation quality.

Furthermore, \METHOD$^\ast$~successfully bridges the efficiency gap. By dynamically adjusting the denoising granularity, \METHOD$^\ast$~increases the average throughput on Dream-Base to 51.26 tokens/s while maintaining a competitive average score of 54.87\%. Notably, on the LLaDA-Inst model for GSM8K, \METHOD$^\ast$~achieves a throughput of 48.02 tokens/s, corresponding to a massive $9.2\times$ speedup over the baseline.

\subsection{Ablation Studies}

We conduct extensive ablation studies to evaluate the contribution of different \METHOD~components. These studies are conducted using the first 200 entries of the GSM8K dataset with Dream-Base.

\textbf{Impact of Decoding Factor}
We compare our parallel decoding approach using different $f$ and $f_r$ (\cref{fig:ablation_f}). To simply, we set $f_r = f / 2.5$. We observe that as $f$ increases, TPS initially exhibits a significant upward trend, reaching its peak at $f=2.5$. However, beyond this point, TPS plateaus and slightly declines and stabilize around $f$ values of $3.0$ and $3.5$. This suggests that while increasing $f$ enhances parallelism and speed up to a certain threshold, the marginal gains in TPS diminish as the rollback overhead begin to outweigh the benefits of a larger $f$.

In contrast, the accuracy remains relatively stable for lower values of $f$, but undergoes a sharp decline once $f$ exceeds $2.5$, dropping to $72.5\%$ at $f=3.5$. Taken together, these results indicate that $f$ around $2.25$ serves as an optimal balance point, achieving a high TPS while maintaining high accuracy.

\textbf{Rollback Budget when Decoding}
We compare our rollback approach using different $R$  (\cref{fig:ablation_R}). Increasing the rollback budget $R$ from $0$ to $2$ leads to a significant improvement in generation accuracy, as it allows the model more opportunities to escape the stagnation trap by revising suboptimal early commitments. However, this gain in quality comes at the expense of throughput, which decreases as more computational resources are allocated to the rollback and re-generation process. Notably, performance begins to plateau or even slightly fluctuate beyond $R=2$, suggesting that a small rollback budget is often sufficient to recover from most early errors while maintaining a favorable efficiency-robustness trade-off.

\subsection{Analysis of Reversibility and Stagnation}

\begin{figure}[tbp]
    \centering
    \includegraphics[width=0.9\linewidth]{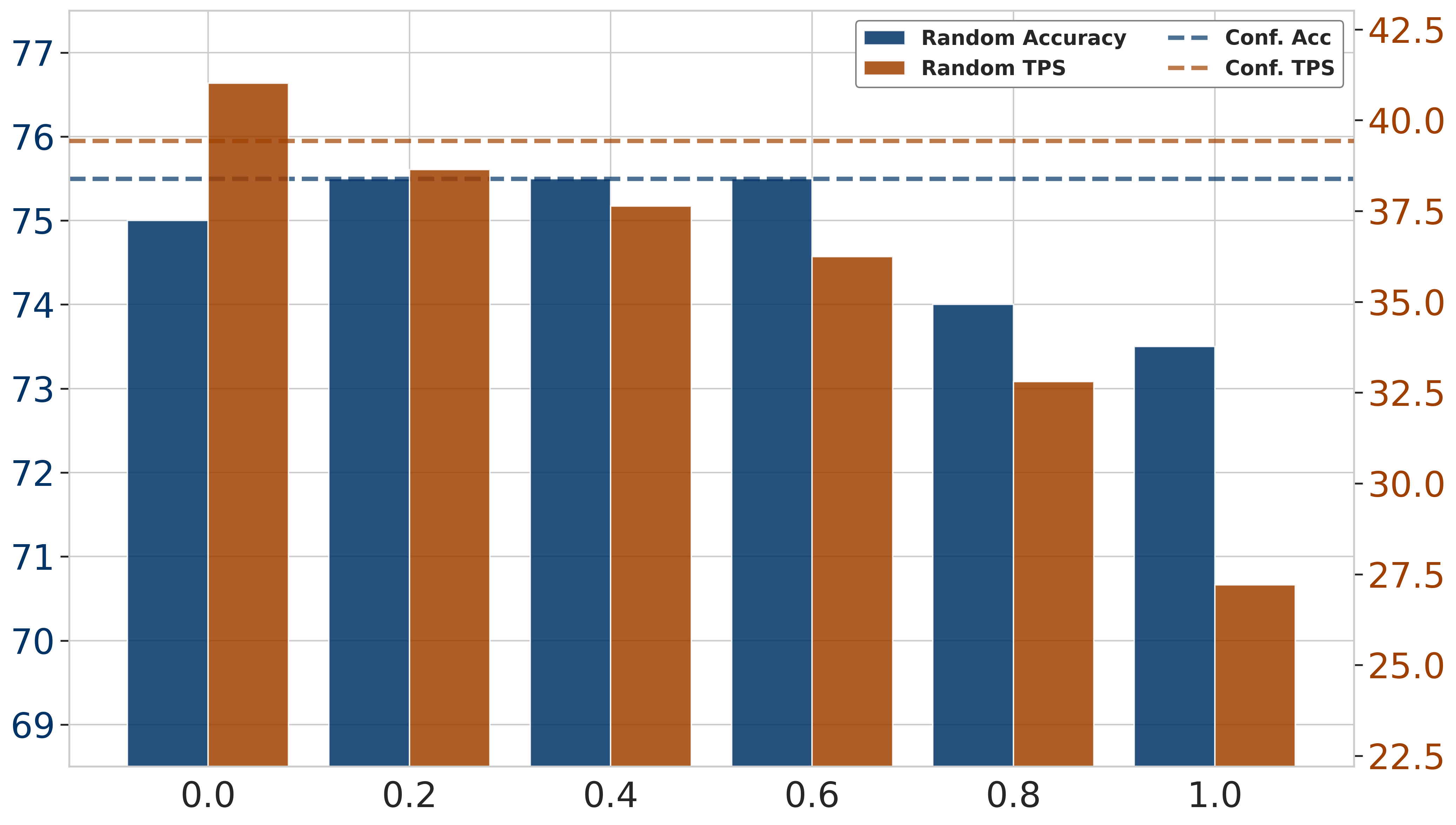}
    \caption{Performance using random remask or confidence-based remask. The x-axis represents the ratio of random remasking.}
    \label{fig:ablation_random}
    \vspace{-19pt}
\end{figure}

\textbf{Random Remask vs. Confidence-Guided Remask}
We evaluate the effectiveness of our confidence-guided remasking by comparing it against a random re-masking baseline with varying ratios. As shown in~\cref{fig:ablation_random}, the confidence-guided strategy consistently outperforms the random remasking approach in terms of accuracy across most ratio settings. While random remasking can achieve competitive accuracy at specific low ratios, its performance and throughput degrade significantly as the remask ratio increases. In contrast, the confidence-guided mechanism strategically targets uncertain tokens for re-initialization based on their individual probability scores rather than a fixed global ratio. This allows~\METHOD~to preserve reliable context more effectively, maintaining a stable and superior balance between generation quality and computational efficiency compared to stochastic methods.

\textbf{Mitigation of Stagnation} 
To quantify the stagnation rate $r_s$ during model generation, we define $r_s$ as the ratio of function evaluations with Fallback $\text{NFE}_f$ to total $\text{NFE}$. We evaluated this metric for Fast-dLLM, \METHOD, and \METHOD$^\ast$ using the first 200 entries of the GSM8K dataset with Dream-Base.

\begin{table}[htbp]
    \centering
    \renewcommand{\arraystretch}{0.95} 
    \caption{Stagnation Comparison: \METHOD~vs. Fast-dLLM}
    \begin{tabular}{lcccc}
        \toprule
        \textbf{Method} & $\text{NFE}_f$ & $\text{NFE}$ & $r_s$ & Score \\
        \midrule
        Fast-dLLM & 8497 & 14508 & 58.57\% & 75 \\
        \METHOD & 9390 & 17118 & 54.85\% & 77 \\
        \METHOD$^\ast$ & 8154 & 15842 & 51.47\% & 76 \\
        \bottomrule
    \end{tabular}
    \label{tab:stagnation_rate}
\end{table}

As presented in~\cref{tab:stagnation_rate}, Fast-dLLM suffers from a high $r_s$ of 58.57\%. This frequent reliance on suboptimal commitments directly constrains its performance, resulting in the lowest accuracy score of 75. In contrast,~\METHOD~successfully reduces the stagnation rate to 54.85\% and achieves the highest overall accuracy of 77. \METHOD~ prevents the accumulation of local errors that otherwise lead to a permanent low-confidence state. \METHOD$^\ast$~achieved a smaller $r_s$ than \METHOD, mainly because \METHOD$^\ast$~used a more relaxed threshold compared to~\METHOD. However, the score of \METHOD$^\ast$~is still higher than Fast-dLLM, demonstrating the effectiveness of our method on reducing $r_s$.







\section{Conclusion}

We identify stagnation as a fundamental failure mode of accelerated diffusion
language model inference, arising from irreversible block-level commitments in
monotonic decoding. To address this issue, we propose \emph{Reversible Diffusion
Decoding (\METHOD)}, a lightweight inference-time framework that enables efficient
rollback and selective re-initialization without retraining or recomputation.
\METHOD~is fully compatible with existing block-wise and cache-aware acceleration
techniques, and can be further enhanced with adaptive scheduling to balance
efficiency and robustness. Experimental results demonstrate that~\METHOD~
significantly improves generation quality and robustness while preserving the
parallel efficiency of diffusion-based decoding.

While \METHOD~effectively mitigates the stagnation trap, it presents certain limitations. First, the method introduces a trade-off between generation quality and throughput; increasing the rollback budget improves accuracy but leads to diminishing returns in speed as computational overhead increases. And the performance of \METHOD~is sensitive to hyperparameters, specifically the scaling factor, requiring careful tuning to find the optimal balance point.

In future work, we plan to explore learning-based policies to dynamically control the rollback and re-masking parameters. Additionally, we aim to investigate \METHOD's applicability to a broader range of long-form text generation tasks beyond reasoning and coding benchmarks.


\section*{Impact Statement}

This paper presents work whose goal is to advance the field of Diffusion Language Model. The datasets and models used in our experiments are all publicly available and widely recognized as standard benchmarks in the field. They do not contain any personal privacy or sensitive information. we acknowledge that improvements in the reliability and speed of large language models carry various potential societal consequences, but none
which we feel must be specifically highlighted here.





\bibliography{References}
\bibliographystyle{icml2026}

\newpage
\appendix
\onecolumn

\section{Stagnation Trap}
\label{app:stagnation-trap}

This section presents an analysis of the \textit{Stagnation Trap}. We provide empirical evidence demonstrating that this phenomenon stems directly from the irreversible accumulation of local errors during monotonic block generation.

\begin{figure}[htbp]
    \centering
    \includegraphics[width=\linewidth]{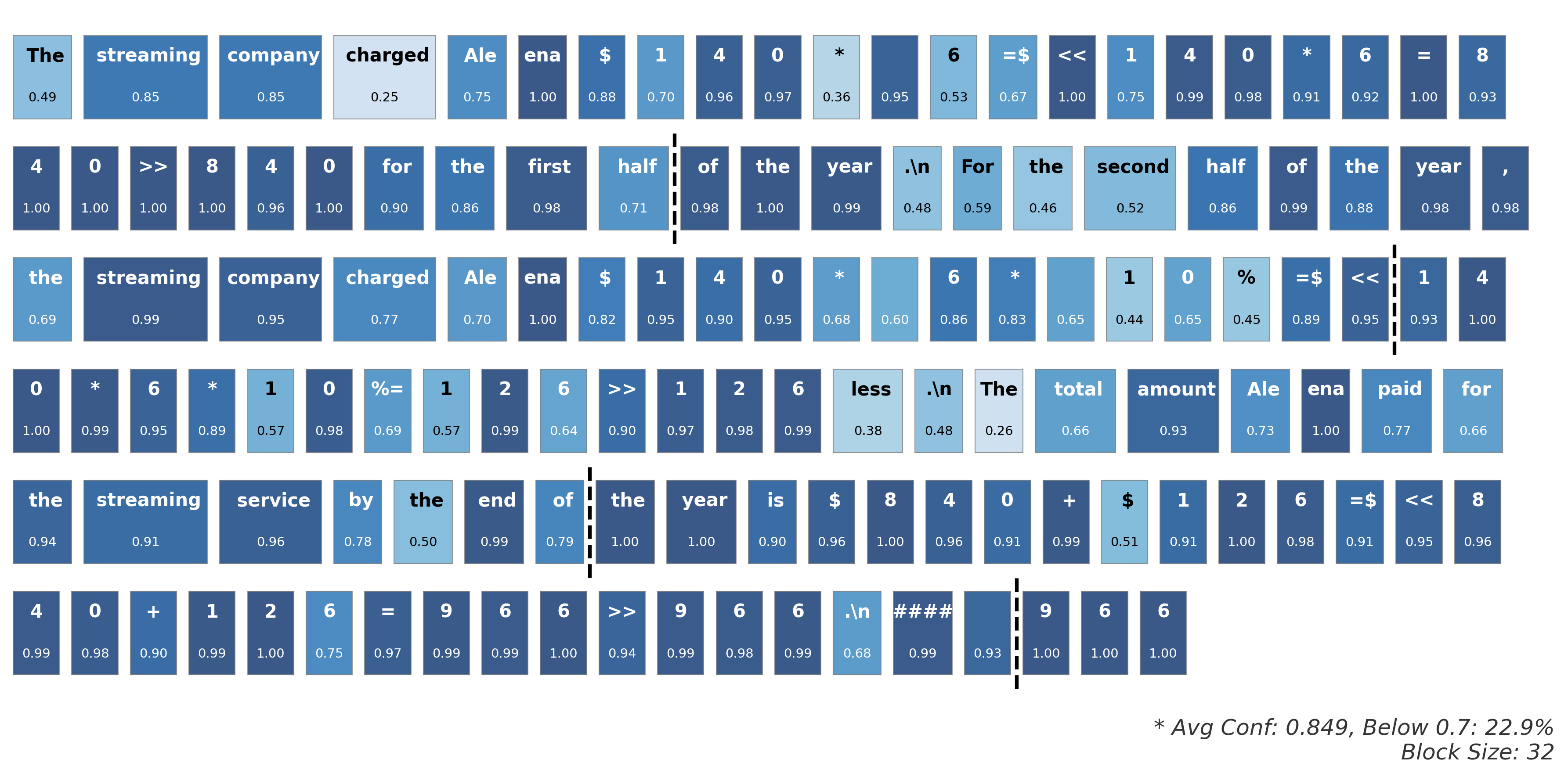}
    \caption{\textbf{Fast-dLLM:} The model relies on fallback mechanisms to resolve uncertainty, leading to low-confidence predictions. Avg Conf: 0.849, with 22.9\% of tokens below 0.7 confidence.}
    \label{fig:app_stagnation_fastdllm}
\end{figure}

\begin{figure}[htbp]
    \centering
    \includegraphics[width=\linewidth]{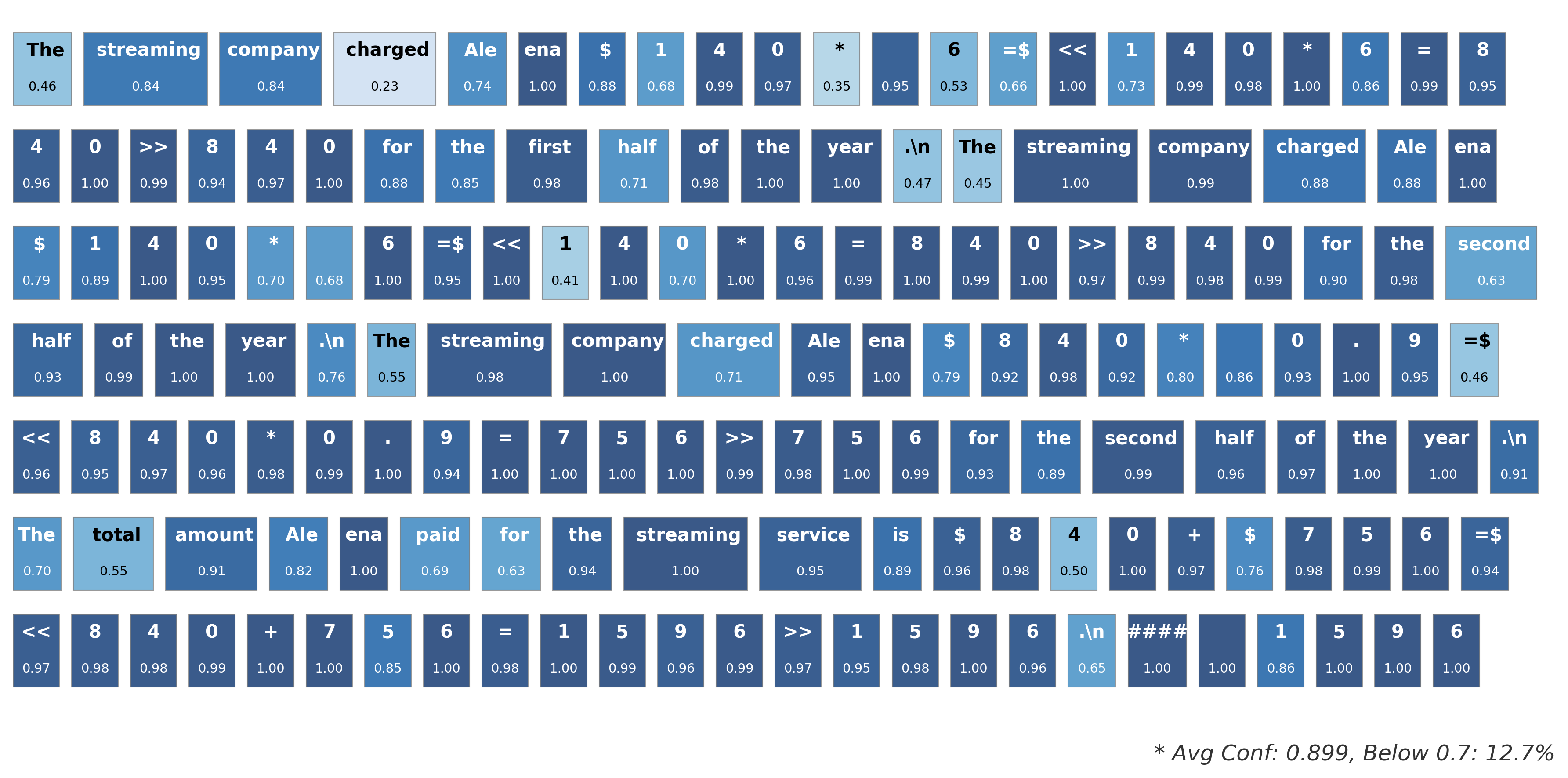}
    \caption{\textbf{\METHOD:} By incorporating rollback, \METHOD~maintains a stable, high-confidence trajectory. Avg Conf: 0.899, with only 12.7\% of tokens below 0.7 confidence.}
    \label{fig:app_stagnation_rdd}
\end{figure}

\begin{figure}[htbp]
    \centering
    \includegraphics[width=\linewidth]{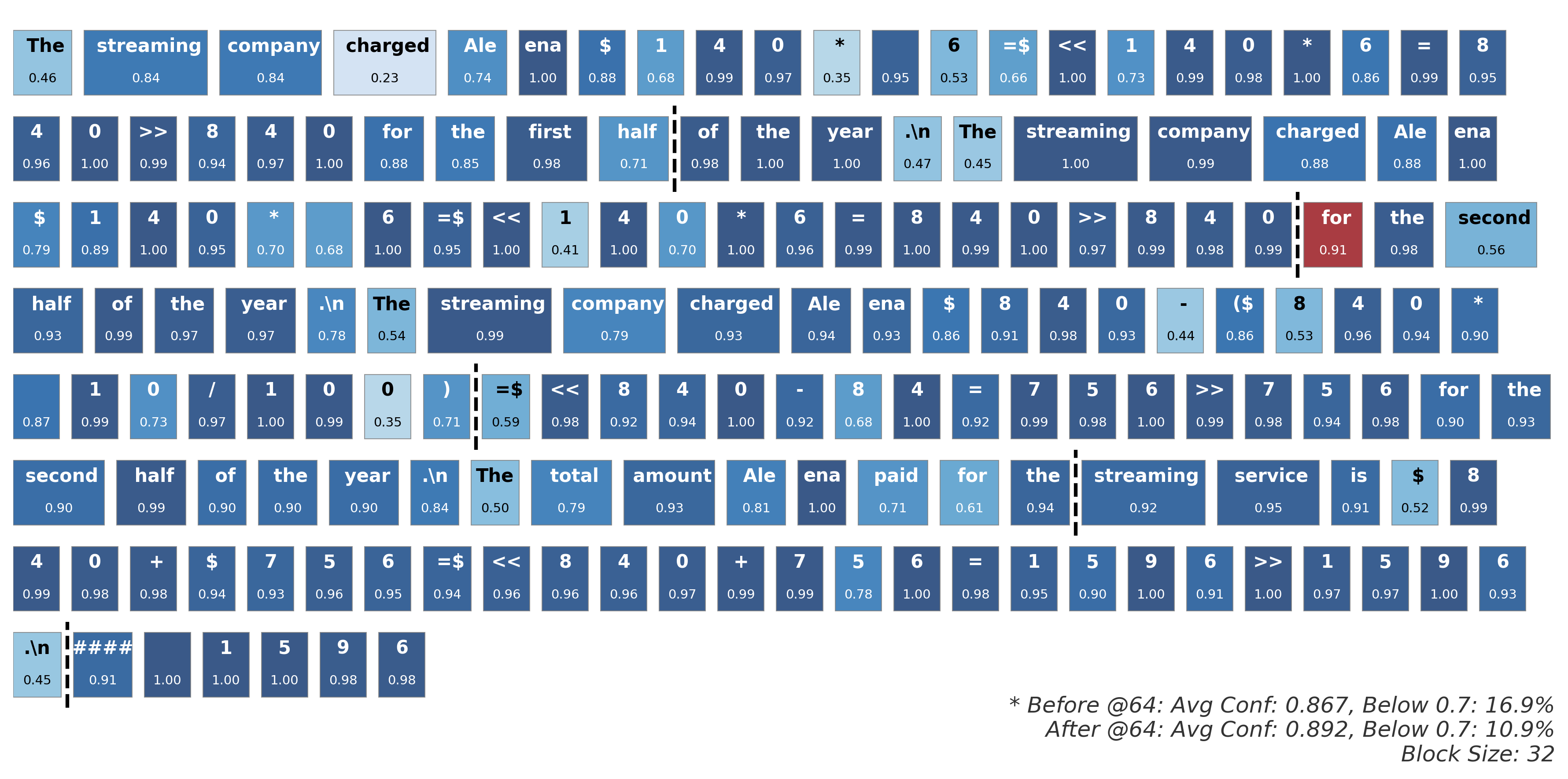}
    \caption{\textbf{\METHOD~as Prefix:} \METHOD~is used to generate the first 64 tokens, followed by Fast-dLLM for the remaining tokens. The high-confidence context provided by \METHOD~(Avg Conf: 0.867) enables Fast-dLLM to sustain high-confidence generation (Avg Conf: 0.892).}
    \label{fig:app_stagnation_rdd_fastdllm}
\end{figure}

\begin{figure}[htbp]
    \centering
    \includegraphics[width=\linewidth]{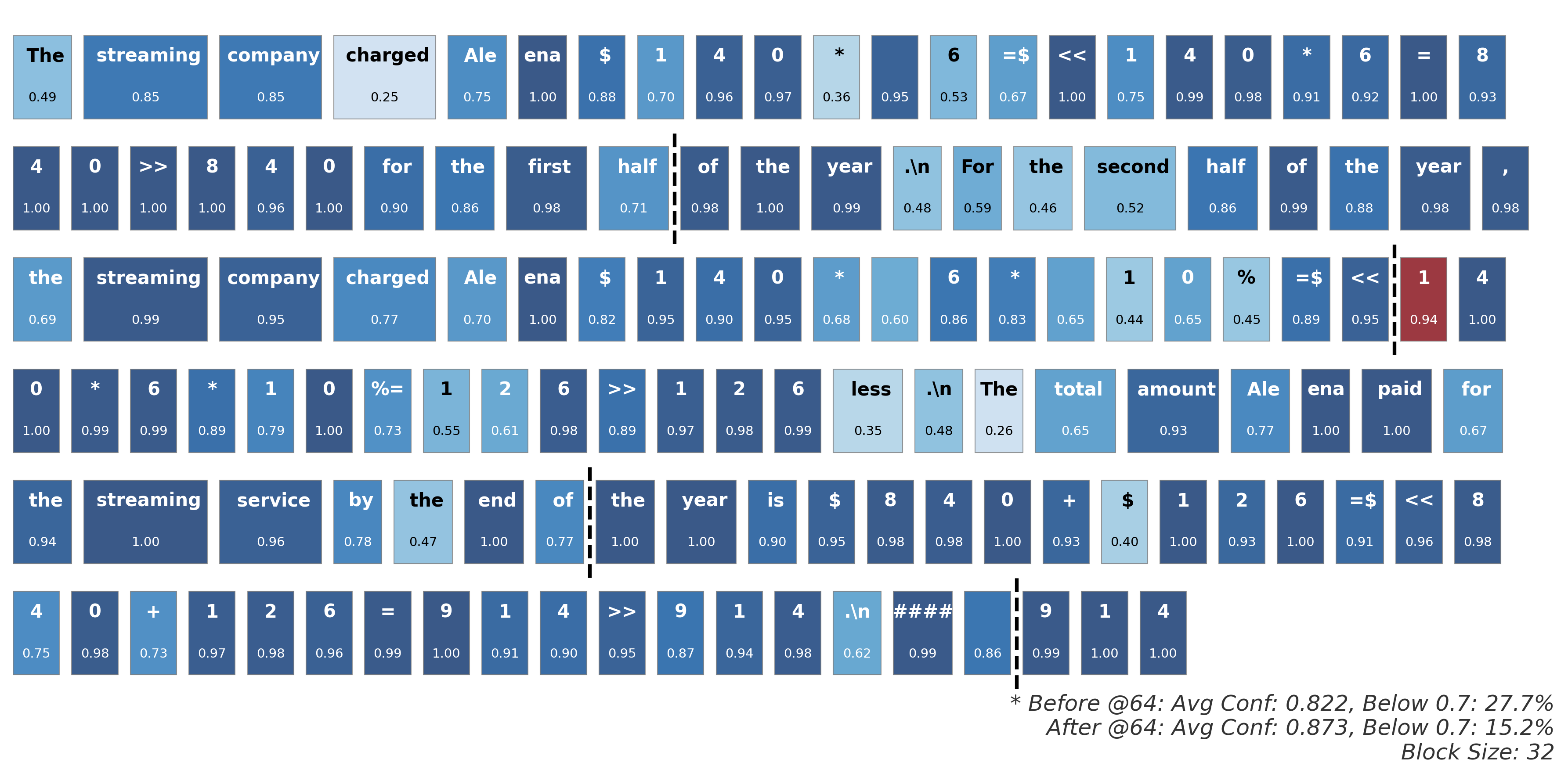}
    \caption{\textbf{Fast-dLLM as Prefix:} Fast-dLLM is used to generate the first 64 tokens, followed by \METHOD. Due to the low-confidence context established by Fast-dLLM (Avg Conf: 0.822), even the robust \METHOD~fails to recover high confidence (Avg Conf: 0.873).}
    \label{fig:app_stagnation_fastdllm_rdd}
\end{figure}

We characterize stagnation as a state where the reverse diffusion process fails to converge to a high-confidence solution within the allocated computational budget. \cref{fig:app_stagnation_fastdllm,fig:app_stagnation_rdd,fig:app_stagnation_fastdllm_rdd,fig:app_stagnation_rdd_fastdllm} contrasts the decoding trajectories of the baseline semi-autoregressive method (Fast-dLLM) and~\METHOD. Darker colors indicate higher confidence.

As illustrated in Figure \ref{fig:app_stagnation_fastdllm}, Fast-dLLM exhibits significant stochastic instability. The prevalence of low-confidence predictions (Avg Conf: $0.849$, with $22.9\%$ of tokens below $0.7$) indicates that the model is conditioned on a suboptimal context. Due to the monotonic constraint, the decoder is forced to accept these marginal predictions, driving the generation into a local optimum from which it cannot escape—the Stagnation Trap. Conversely, Figure \ref{fig:app_stagnation_rdd} demonstrates the efficacy of the rollback mechanism.~\METHOD~maintains a consistently high-confidence trajectory (Avg Conf: $0.899$). By detecting stagnation onset and performing targeted re-masking, the method effectively prunes low-probability branches before they solidify into the context, thereby preserving global coherence.

To verify that stagnation is a state-dependent failure rather than an inherent model limitation, we conducted an intervention study by switching decoding strategies at a fixed checkpoint (token 64). This experiment isolates the impact of prefix quality on subsequent generation stability, as shown in~\cref{fig:app_stagnation_fastdllm_rdd,fig:app_stagnation_rdd_fastdllm}.

In~\cref{fig:app_stagnation_rdd_fastdllm}, the sequence is initialized using~\METHOD, generating a high-confidence prefix (Avg Conf: $0.867$). We observe that switching to the standard Fast-dLLM for the suffix does not degrade performance; in fact, the baseline model achieves a high average confidence of $0.892$, close to the \METHOD's 0.899. This result suggests that standard block diffusion is capable of high-quality generation provided the conditioning context remains within the manifold of valid text. However, in~\cref{fig:app_stagnation_fastdllm_rdd}, we initialize with Fast-dLLM, resulting in a degraded prefix (Avg Conf: $0.822$). Crucially, even when switching to the more capable~\METHOD~algorithm, the generation fails to reach the confidence levels seen in the pure~\METHOD~setting.


\end{document}